\documentclass[conference]{IEEEtran}
\makeatletter
\def\footnoterule{\relax
  \kern-5pt
  \hbox to \columnwidth{\hfill\vrule width 0.8\columnwidth height 0.4pt\hfill}
  \kern4.6pt}
\makeatother
\usepackage{cite}
\ifCLASSINFOpdf
   \usepackage[pdftex]{graphicx}
   \graphicspath{{../pdf/}{../jpeg/}}
   \DeclareGraphicsExtensions{.pdf,.jpeg,.png}
\else
   \usepackage[dvips]{graphicx}
   \graphicspath{{../eps/}}
   \DeclareGraphicsExtensions{.eps}
\fi
\usepackage{siunitx}
\usepackage{multirow}
\usepackage{xcolor}
\definecolor{redcolor}{rgb}{1, 0, 0}
\usepackage{subfiles}
\usepackage{makecell}
\usepackage{color,soul}
\usepackage{amsmath}

\IEEEoverridecommandlockouts

\usepackage [english]{babel}
\usepackage [autostyle, english = american]{csquotes}
\MakeOuterQuote{"}

\newcommand\blfootnote[1]{%
  \begingroup
  \renewcommand\thefootnote{}\footnote{#1}%
  \addtocounter{footnote}{-1}%
  \endgroup
}

\title{\vspace{0.8in} Passive frustrated nanomagnet reservoir computing}
\author {
    \vspace{0.15in} \parbox{\linewidth}{\textbf{Alexander J. Edwards\textsuperscript{1*}, Dhritiman Bhattacharya\textsuperscript{2}, Peng Zhou\textsuperscript{1}, Nathan R. McDonald\textsuperscript{3}, Walid Al Misba\textsuperscript{2}, Lisa Loomis\textsuperscript{3}, Felipe Garc\'ia-S\'anchez\textsuperscript{4}, Naimul Hassan\textsuperscript{1}, Xuan Hu\textsuperscript{1}, Md. Fahim Chowdhury\textsuperscript{2}, Clare D. Thiem\textsuperscript{3}, Jayasimha Atulasimha\textsuperscript{2}, Joseph S. Friedman\textsuperscript{1*}}}%
}

\begin{document}

\twocolumn[
  \begin{@twocolumnfalse}

\maketitle

\begin{abstract}
\vspace{-0in}\textbf{Reservoir computing (RC) has received recent interest because reservoir weights do not need to be trained, enabling extremely low-resource consumption implementations, which could have a transformative impact on edge computing and in-situ learning where resources are severely constrained.  Ideally, a natural hardware reservoir should be passive, minimal, expressive, and feasible; to date, proposed hardware reservoirs have had difficulty meeting all of these criteria.  We therefore propose a reservoir that meets all of these criteria by leveraging the passive interactions of dipole-coupled, frustrated nanomagnets.  The frustration significantly increases the number of stable reservoir states, enriching reservoir dynamics, and as such these frustrated nanomagnets fulfill all of the criteria for a natural hardware reservoir. We likewise propose a complete frustrated nanomagnet reservoir computing (NMRC) system with low-power complementary metal-oxide semiconductor (CMOS) circuitry to interface with the reservoir, and initial experimental results demonstrate the reservoir's feasibility.  The reservoir is verified with micromagnetic simulations on three separate tasks demonstrating expressivity. The proposed system is compared with a CMOS echo-state-network (ESN), demonstrating an overall resource decrease by a factor of over 10,000,000, demonstrating that because NMRC is naturally passive and minimal it has the potential to be extremely resource efficient.}\vspace{0.4in}

\end{abstract}

\vspace{-2em}
  \end{@twocolumnfalse}
]

\begin{figure*}
    \centering
    \includegraphics[width=\textwidth]{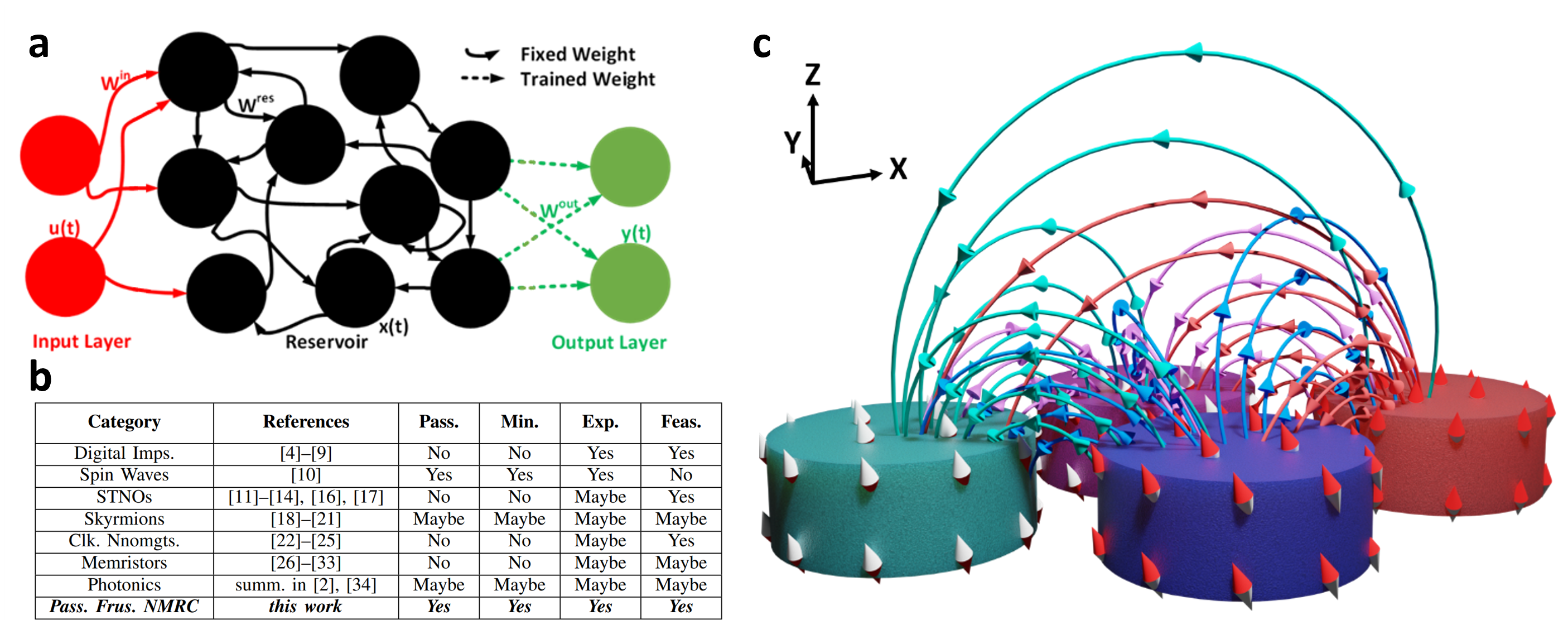}
    \caption{\textbf{a}, A reservoir computer is an RNN with untrained input and hidden layer weights.  A single trained linear output layer is sufficient to extract a meaningful task output.  \textbf{b}, Previous hardware RC proposals and their shortcomings. \textbf{c}, Frustrated perpendicular magnetic anisotropy (PMA) nanomagnets.  PMA nanomagnets naturally relax in the perpendicular direction ($\pm z$).  Compass needles depict magnetization direction.  The colored arrows depict a small portion of the magnetic field lines produced by that color of nanomagnet upon the other nanomagnets. (The lower half of the field lines have been omitted for space.)  These nanomagnets are frustrated because they cannot all rest in their isolated lowest energy states but must instead find an equilibrium lowest energy state.\vspace{-1em}}
    \label{fig:Figure1_RC_Frus}
\end{figure*}

\blfootnote{\textsuperscript{1}Department of Electrical and Computer Engineering, The University of Texas at Dallas, Richardson, TX \textsuperscript{2}Department of Mechanical and Nuclear Engineering, Virginia Commonwealth University, Richmond, VA \textsuperscript{3}Air Force Research Laboratory - Information Directorate, Rome, NY \textsuperscript{4}Universidad de Salamanca, Salamanca, Spain \textsuperscript{*}e-mail: Joseph.Friedman@utdallas.edu; Alexander.Edwards@utdallas.edu $\vert$ Approved for Public Release; Distribution Unlimited: AFRL-2022-4420.}
As training an artificial neural network has enormous hardware costs, reservoir computers (RCs) are promising due to their ability to provide accuracies similar to trained recurrent neural networks (RNNs) while requiring only minimal training \cite{Jaeger_Haas2004_RC, recentAdvancesHirose}. A reservoir projects input stimuli into a higher dimensional space, enabling linearly-separable solutions to complex temporal problems.  As illustrated in Fig. \ref{fig:Figure1_RC_Frus}a, reservoir weights are untrained, circumventing the costs of training programmable weights; fixed physical structures can be used that do not incur the area, energy, and delay costs inherent to programmability.  An early conceptualization demonstrated computation via a reservoir of water by extracting the interactions among ripples in the water \cite{WaterReservoir-2003}; while slow and large, the water reservoir exemplifies the ability of hardware reservoirs to naturally process information through elegant physical interactions.
    
As summarized in Fig. \ref{fig:Figure1_RC_Frus}b, none of the previously proposed hardware reservoirs exhibit all of the characteristics of an ideal reservoir:
\begin{itemize}
    \item Passive: Use natural physical interactions with energy only provided to the reservoir through input information.
    \item Minimal: Naturally process information without requiring complex external circuitry or post-processing.
    \item Expressive: Demonstrate high dimensionality, nonlinearity, fading memory, and the separation property \cite{recentAdvancesHirose}. 
    \item Feasible: Facile system fabrication that considers input/output interfacing, noise, scalability, and timing.
\end{itemize}
    
\noindent While hardware reservoirs implemented in digital logic \cite{ECARC_Snyder13, ECARC_McDonald17, ECARC_moran18,RNN_ESN_Honda20, RNN_ESN_Liao20,StchRC_Alomar16} are highly expressive and clearly feasible, they are neither passive nor minimal. The most promising spintronic reservoir is based on spin-waves \cite{spin-wave-garnet-film_Nakane2018}, though the difficulty in probing the spin-wave states impedes practical system demonstrations. Other spintronic proposals utilize spin torque nano-oscillators with constant bias currents \cite{STNOs_Furuta2018, STNOs_Markovic2019, STNOs_Riou2019, STNOs_Tsunegi2019, STNOs_Kanao2019, STNOs_Yamaguchi2020a, STNOs_Yamaguchi2020b} or that require additional circuitry for post-processing before feeding into the output layer \cite{STNOs_Riou2019, STNOs_Yamaguchi2020a, STNOs_Tsunegi2019}, while skyrmion reservoir proposals are insufficiently mature for reliable characterization \cite{skyrmion_reservoirs_Prychynenko2018, skyrmion_reservoirs_Pinna2020, skyrmion_reservoirs_EverschorSitte21, AtulSkyrmRC}.  Another spintronic reservoir proposal uses arrays of nanomagnets which are actively clocked \cite{clked_nms_Nomura2019, clked_nms_Nomura2020, clocked_nms_nomura2021, ClockedNMs_Hon21}, making these reservoirs neither passive nor minimal. Reservoirs with non-volatile memristors \cite{memristor_networks_Bennett2017, memristor_networks_Burger2013, memristor_networks_Burger2015, memristor_networks_Du2017, memristor_networks_Hassan2017, memristor_networks_Kulkarni2012, memristor_networks_Tanaka2017, memristor_networks_Zhong2021} must be actively pulsed to relax the state, while photonic reservoirs \cite{recentAdvancesHirose, PhotonicRC_review_Vandersande17} process information too fast relative to the hysteretic memory time constants or require additional circuitry \cite{PhotonicRC_mask_Shneider16, PhotonicRC_TDR_Antonik17} to maintain the information in the reservoir.
    
We therefore leverage frustrated nanomagnetism in a proposal for the first hardware RC that is passive, minimal, expressive, and feasible. Frustration between nanomagnets enables efficient and complex computation through the passive relaxation of the coupled magnetizations.  We describe a complete nanomagnet reservoir computing (NMRC) system with inherent passivity and minimality, and provide an initial experimental demonstration of its feasibility. The expressivity of this frustrated nanomagnet reservoir was demonstrated through simulations for three benchmark tasks, and evaluation of hardware resource usage indicates a decrease by a factor of over ten million as compared to CMOS.
    
\vspace{\baselineskip}
{\large \noindent \textbf{Reservoir computing with frustrated \\nanomagnets}}

We propose an energy-efficient hardware reservoir in which passive coupling among frustrated nanomagnets generates high-dimensional information processing. As illustrated in Fig. \ref{fig:Figure1_RC_Frus}c, each individual nanomagnet naturally relaxes along its easy axis -- which in this work is the out-of-plane $z-$axis -- resulting in two stable magnetization states. When nanomagnets are packed closely in an irregular array, they can become frustrated; that is, their coupling prevents some or all of the magnets from relaxing along an easy axis. The system's relaxation to a global minimum energy does not result in the minimization of each nanomagnet's energy. This nanomagnetic frustration enables the system to relax to a broad range of local energy minima, producing the critical RC features of large expressivity and hysteretic memory. As opposed to regular lattice arrangements \cite{SpinIce-Jensen20, clked_nms_Nomura2019, clked_nms_Nomura2020, clocked_nms_nomura2021, ClockedNMs_Hon21}, the reservoirs are designed with an irregular asymmetric layout that enriches the reservoir expressivity.

\begin{figure*}
    \centering
    \includegraphics[width=\textwidth]{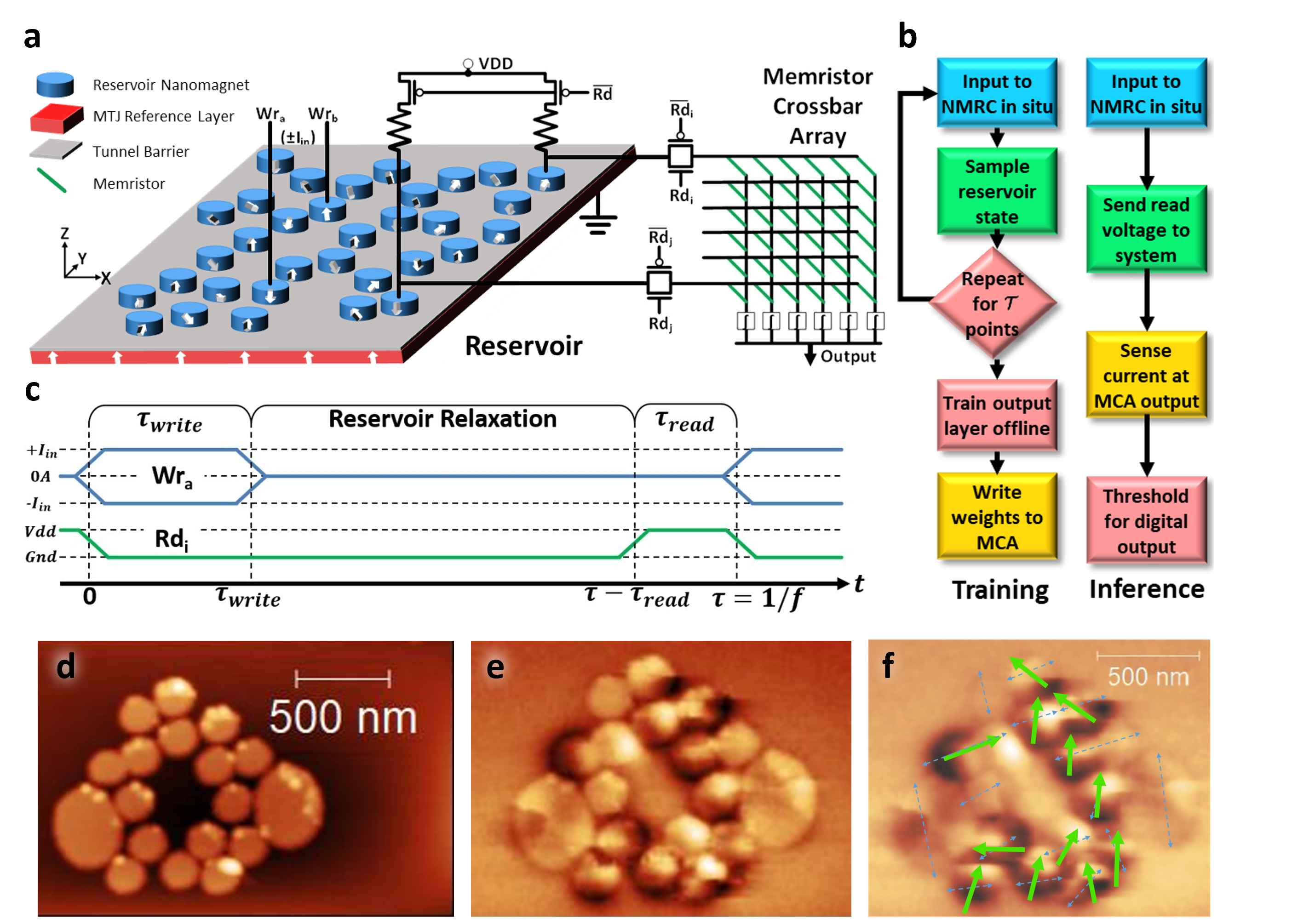}
    \caption{\textbf{a}, NMRC system diagram.  \textbf{b}, NMRC training and inference process.  Blue boxes represent an input to the reservoir via the $Wr$ signals.  Green boxes represent a read to the reservoir via the $Rd$ signals.  Red boxes represent an action performed outside of the proposed system, and yellow boxes represent a step performed by the MCA.  \textbf{c}, NMRC inference timing.  \textbf{d}, topography image showing a planar array of dipole coupled nanomagnets where the larger magnets can act as inputs. \textbf{e}, phase image showing the magnetic contrast. \textbf{f}, illustration of the magnetization direction (solid green arrows) and easy axis (dashed blue arrows) of the individual magnets based on the topography and the phase images, demonstrating a frustrated magnetic state.}
    \label{fig:Figure2_NMRC}
\end{figure*}
    
This nanomagnet reservoir can be readily integrated with conventional technologies in a complete system as illustrated in Fig. \ref{fig:Figure2_NMRC}a. The input signals can be provided through spin-transfer torque (STT) \cite{STT_Tutorial08} or spin-orbit torque (SOT) \cite{Lee2018_SOT} switching of the nanomagnets, with binary input signaling due to the bistable nature of these nanomagnets. In response to this input, the reservoir passively relaxes toward an energy minimum through an exploration of a rich, temporally evolving landscape of magnetizations. To read the information in the nanomagnet reservoir, magnetic tunnel junctions (MTJs) (the central components of magnetoresistive random access memory (MRAM) \cite{MRAM-VLSI}) can be formed by patterning the reservoir nanomagnets atop a tunnel barrier and pinned ferromagnetic layer to enable the magnetization state to be determined through a voltage divider.

Whereas the nanomagnet reservoir has fixed couplings that can be considered to represent synaptic weights, the single-layer memristor crossbar array (MCA) is trained in a supervised manner. In particular, the MCA performs the vector-matrix multiplication  \cite{MCA_review_Li18, MCA_Hu18, MCA_Cai19, MCA_Chen19} to produce the system output
\begin{equation}
    \mathbf{\hat{y}} = \mathbf{W^{out}} * \mathbf{X},
    \label{eq1}
\end{equation}
where $\mathbf{X}$ is the reservoir voltage landscape and $\mathbf{W^{out}}$ is the trained output-weight vector (see Methods). This system is restricted to binary outputs to circumvent the need for an analog-to-digital converter (ADC), thereby minimizing hardware costs.  All of these technologies are available in modern lithographic processes and have been experimentally demonstrated, providing a feasible path to production for an integrated NMRC.  In fact, we have fabricated an initial proof-of-concept reservoir layer of in-plane nanomagnets (Fig. \ref{fig:Figure2_NMRC}d-f) exhibiting frustrated magnetization states (see Methods) and demonstrating that the nanomagnet reservoir is feasible and has an open path to production with modern processes.

The training and inference processes are depicted in Fig. \ref{fig:Figure2_NMRC}b, with a single inference operation illustrated in Fig. \ref{fig:Figure2_NMRC}c. Each input signal is provided through a write pulse that forces input nanomagnets to a fixed state, causing the reservoir to reach a new state that is a function of the input and the previous state (and therefore, the past inputs). To read the reservoir state $\mathbf{X}$, voltage pulses are provided to the output MTJs, which directly drive the MCA to produce a task output.  The reservoir is minimal and passive, as it only requires energy to be applied through the input information, thereby enabling extremely low-power computation.


\begin{figure*}
    \centering
    \includegraphics[width=\textwidth]{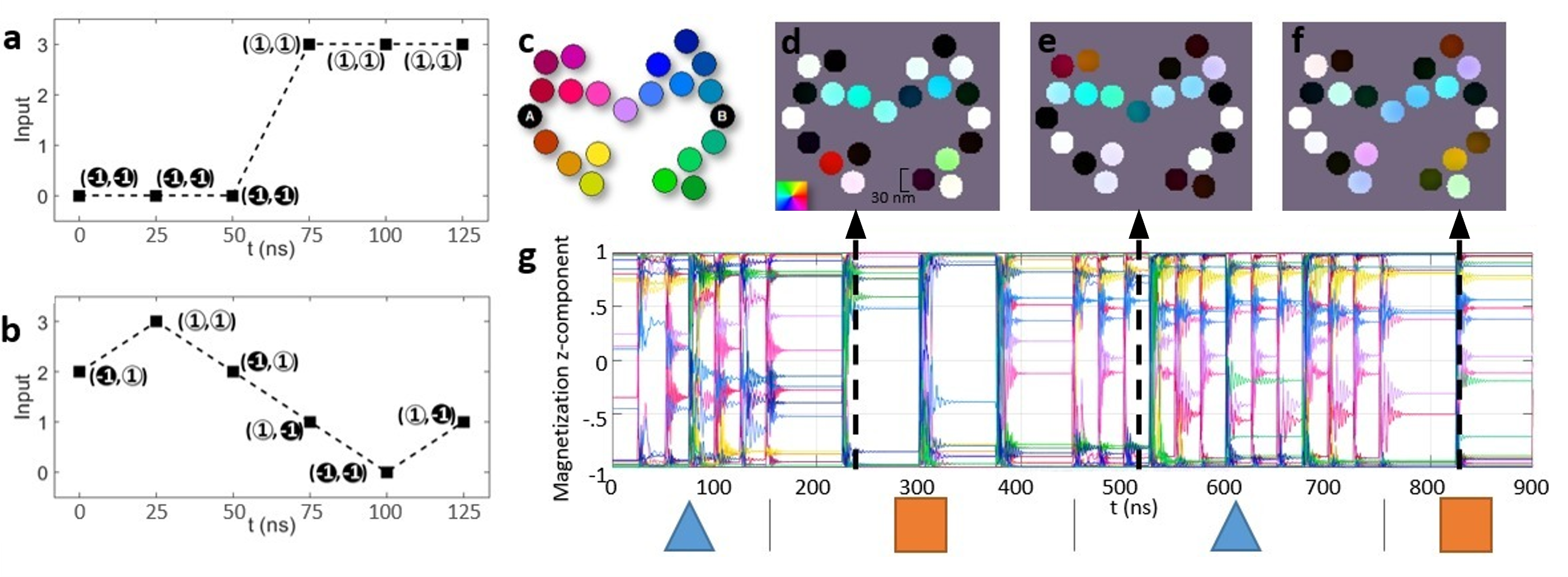}
    \caption{\textbf{Triangle and square waveform identification task.} \textbf{a-b,} Square-wave and triangle-wave input sequences, respectively, are randomly concatenated.  White (black) inputs represent magnetization in the $+z$ ($-z$) direction. \textbf{c,} Reservoir layout, with input nanomagnets $A$ and $B$ driving the propagation of information through the colored nanomagnets. \textbf{d-f,} Simulation snapshots. As noted in the color wheel inset, white (black) represents magnetization along the $+z$ ($-z$) axis while the colors of the rainbow represent magnetizations in the $xy$ plane.  \textbf{g,} Evolution of the magnetization z-components during a portion of the simulation, with the colors of each line matching the nanomagnet colors in \textbf{c}.  Shapes below the traces indicate the input waveform, while dashed lines pointing to \textbf{d-f} indicate when these snapshots were taken.}
    \label{fig:Figure3_SinSqu}
\end{figure*}
    
    \vspace{\baselineskip}
    
{\large \noindent \textbf{Nanomagnet reservoir information processing}}

To evaluate the ability of frustrated nanomagnet reservoirs to support high-dimensional short-term memory, micromagnetic simulations \cite{mumax3} were performed on three benchmark RC tasks (see Methods). The nanomagnet reservoir performance was compared against an RC without a reservoir layer, in which a single linear output layer with no reservoir is presented with delayed copies of the task inputs (see Methods); this network is equivalent to a single-layer perceptron network or linear classifier. For all three tasks, the NMRC achieved a significantly higher accuracy than the RC with no reservoir layer, demonstrating that passive nanomagnet reservoirs can perform complex, non-linear, temporal functions with high expressivity.

\vspace{\baselineskip}
{\noindent \textit{Triangle-square wave identification}}
    
The waveform identification task is a common simple benchmark task for reservoirs \cite{memristor_networks_Kulkarni2012, memristor_networks_Burger2013, memristor_networks_Tanaka2017,  skyrmion_reservoirs_Pinna2020, memristor_networks_Zhong2021}, requiring the reservoir to differentiate between triangle and square waveforms presented through time.  As the NMRC operates on binary inputs, triangle and square waves were quantized to two bits before being input to the reservoir as shown in Fig. \ref{fig:Figure3_SinSqu}. After training, the NMRC achieved $100\%$ classification accuracy on input waveforms from a testing data set.  The RC with no reservoir only obtained $79\%$ accuracy, demonstrating that the frustrated nanomagnets exhibit high expressivity. 
    
\vspace{\baselineskip}
{\noindent \textit{Boolean function evaluation}}
    
Two of the most widely used metrics for RC are short-term-memory (STM) and parity-check (PC) \cite{STNOs_Furuta2018, STNOs_Tsunegi2019, STNOs_Kanao2019, STNOs_Yamaguchi2020a, STNOs_Yamaguchi2020b, ClockedNMs_Hon21}, which require the reservoir to, respectively, remember the previous $k$ inputs of an input bitstream or to perform $k$-bit XOR on those bits.  This provides gauges for the memory content (STM) and non-linear expressivity (PC).  The Boolean function evaluation task is a superset of the STM and PC tasks, requiring the RC to perform arbitrary $k$-bit Boolean functions including STM and PC.  Accuracies for each of the $2^{2^k}$ $k$-bit functions were averaged together to calculate an overall metric for reservoir accuracy.

\begin{figure*}
    \centering
    \includegraphics[width=\textwidth]{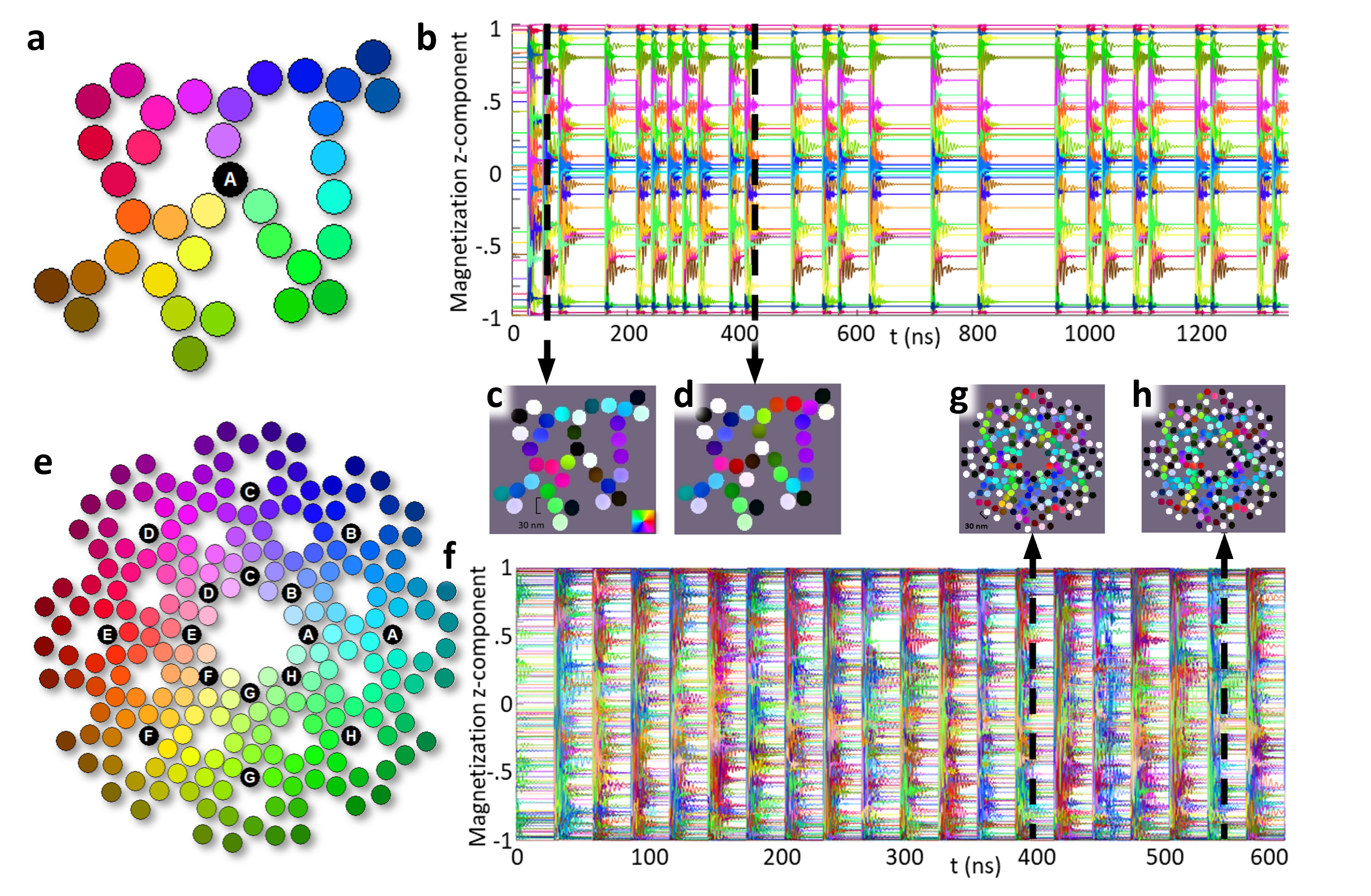}
    \caption{\textbf{Sophisticated RC tasks.} \textbf{a-d}, Boolean function evaluation task: \textbf{a,} Reservoir layout. The input nanomagnet is depicted as black with white text. \textbf{b}, Evolution of the magnetization z-components during a portion of the simulation, with the colors of each line matching the nanomagnet colors in \textbf{a}. Dashed lines pointing to \textbf{c, d} indicate when these snapshots were taken. \textbf{c, d}, Simulation snapshots.  \textbf{e-h}, ECA observer with $k = 4$. \textbf{e}, Reservoir layout.  \textbf{f}, Magnetization z-components during a portion of the simulation. \textbf{g, h}, Simulation snapshots. }
    \label{fig:Figure4_Tasks}
\end{figure*}

    
The nanomagnet reservoir illustrated in Fig. \ref{fig:Figure4_Tasks}a-d performed the Boolean function evaluation task with $100\%$ accuracy for both $k = 2$ and $3$ bits, and $93.4\%$ accuracy for $k = 4$.  In contrast, the RC with no reservoir layer performed the Boolean function evaluation task with $99.6\%$ accuracy for $k=2$, $81.4\%$ for $k=3$, and $84.7\%$ for $k=4$. Furthermore, the nanomagnet reservoir achieved an STM content of $4.68$ bits, which is standard for spintronic reservoirs, and a PC capability of $3.73$ bits, which is close to the maximum PC reported in the literature for emerging hardware reservoirs \cite{STNOs_Furuta2018, STNOs_Yamaguchi2020a, STNOs_Yamaguchi2020b, STNOs_Tsunegi2019, STNOs_Kanao2019, ClockedNMs_Hon21} (see Methods). These results further demonstrate the ability of the NMRC to perform high-dimensional information processing with large expressivity and memory content.
    
\vspace{\baselineskip}
    
{\noindent \textit{Elementary cellular automata observer}}
    
Observer tasks predict the internal state of a highly complex dynamical system given only an observed state, and have therefore received significant interest as an appropriate mapping to RC \cite{RCobservers}.  Elementary cellular automata (ECA) have demonstrated a wide range of dynamics \cite{wolframECA} making them a suitable candidate for a discrete-time binary system to observe.  An ECA grid is generated with an arbitrary first row, with each successive row generated by applying ECA rule 59 to the previous row, wrapping at the edges.  Rule 59 was chosen because of its complex yet predictable, periodic behavior.  The reservoir must reproduce the entire ECA grid after receiving data from eight evenly spaced input columns from this grid provided sequentially row-by-row.  The spacing between successive input columns is denoted by $k$, and the entire grid has width $8k$.  For the relatively simple case of $k = 4$, every fourth column of the $32$-column-wide ECA grid is input to the reservoir, which must reproduce the $24$ unknown columns along with the eight known columns.
      
\begin{figure*}
    \centering
    \includegraphics[width=\textwidth]{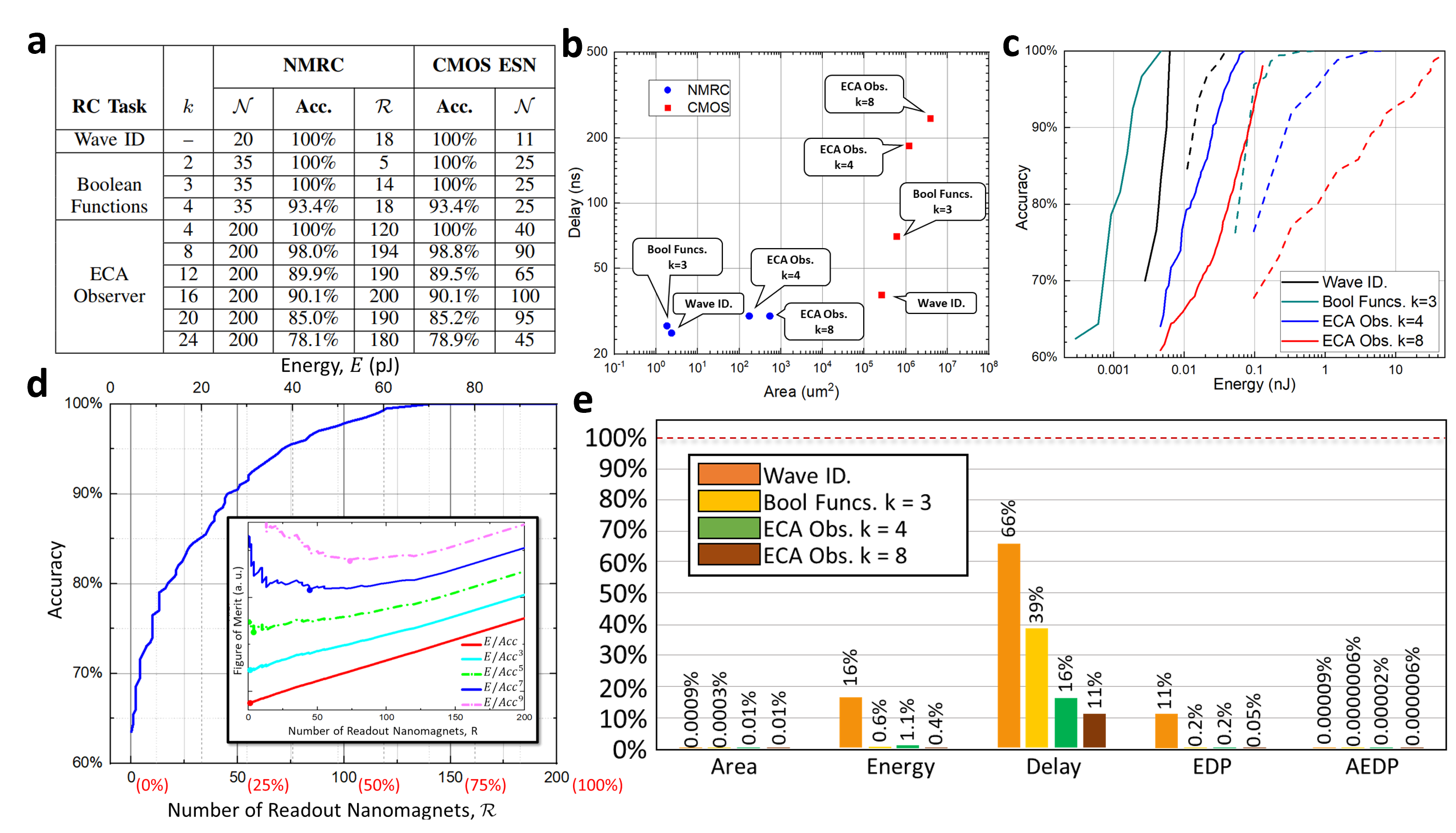}
    \caption{\textbf{NMRC results} \textbf{a}, NMRC accuracy for each task compared with equivalent CMOS ESN.  \textbf{b}, Scatter plot showing delay and area for NMRC and CMOS for various tasks.  \textbf{c}, Energy vs. accuracy for all three tasks with various values of $k$ when implemented in CMOS and NMRC.  Solid (dashed) lines correspond with the NMRC (CMOS) system.  The energy decrease between NMRC and a CMOS reservoir increases with increasing task complexity, indicating that NMRC scales better than CMOS for complex tasks. \textbf{d}, Trade-off between accuracy and energy costs by varying the number of readout magnets, $\mathcal{R}$, among the $200$ reservoir magnets for the ECA observer with $k = 4$. Inset: In resource-constrained contexts, the accuracy and energy trade-off may be optimized through the minimization of an application-dependent metric function; five such functions are shown. \textbf{e}, Area, energy, delay, energy-delay product (EDP), and AEDP of NMRC normalized according to the CMOS benchmark indicated with the dashed line at $100\%$.  NMRC is $44,000$ times more area-efficient,  $60$ times more energy-efficient, and four times more time-efficient, giving a combined factor of $240$ for EDP, and $10,000,000$ for AEDP.    \vspace{1em}}
    \label{fig:Figure5_Results}
\end{figure*}

The $200$-nanomagnet NMRC in Fig. \ref{fig:Figure4_Tasks}e-h has a circular structure that matches the edge-wrapping of the ECA grid, with each input duplicated through two nanomagnets for increased expressivity.  The reservoir attained $100\%$ accuracy for $k = 4$, $98.2\%$ for $k = 8$, decreasing with increasing $k$ to provide $78.1\%$ accuracy for $k = 24$; the full table of accuracies is displayed in Fig. \ref{fig:Figure5_Results}a.
Due to the periodicity of ECA rule 59, the RC with no reservoir layer was restricted in memory to evaluate the reservoir's expressivity, and it therefore achieved only $91.8\%$ accuracy for $k=4$, $79.1\%$ for $k=8$, and decreased to $73.0\%$ for $k = 24$.  The nanomagnet reservoir thus achieved higher accuracies than the RC with no reservoir layer for all values of $k$, indicating that the reservoir is performing expressive, non-linear computation.

\vspace{\baselineskip}

{\large \noindent \textbf{Computing efficiency and outlook}}

To estimate the advantages provided by NMRC in terms of area, energy, and delay, NMRC was compared to an equivalent CMOS echo state network (ESN) (see Methods). To ensure a fair comparison with equivalent accuracies, numerous CMOS ESNs were designed (see Methods) with varying numbers of neurons to achieve the same accuracies as the NMRC for each task (Fig. \ref{fig:Figure5_Results}a) . As both the CMOS RC and NMRC require an MCA output layer, identical assumptions were made regarding memristor parameters.

NMRC provides massive efficiency advantages over CMOS RC in terms of area, speed, and energy. As shown in Fig. \ref{fig:Figure5_Results}b, NMRC has smaller area and a delay that is a nearly constant function of task complexity, whereas the CMOS ESN needs significantly more time or area to provide equivalent accuracy. NMRC consumes significantly less energy, as can be readily observed in Fig. \ref{fig:Figure5_Results}c. Importantly, a large proportion of the reservoir nanomagnets can passively contribute to the computation without being actively used as output, enabling drastic energy savings as shown in Fig. \ref{fig:Figure5_Results}d.

Overall, as illustrated in Fig. \ref{fig:Figure5_Results}e, NMRC provides a reduction in area by a factor of $44,000$, in energy by a factor of $60$, and in delay by a factor of four, culminating in the improvement of area-energy-delay product (AEDP) by a factor greater than $10,000,000$. While these results do not consider fabrication imprecision or stray fields from the MTJ reference layer and were achieved at zero temperature, promising engineering solutions exist for all of these challenges (see Methods); furthermore, the irregularity necessary for this system makes it inherently robust against concerns related to fabrication imprecision. The passivity and minimality of NMRC, coupled with its expressivity and feasibility, therefore provides a promising solution for artificial intelligence applications with extreme efficiency.


\vspace{\baselineskip}
    
{\noindent \large  \textbf{Methods}}
\footnotesize
    
 \vspace{\baselineskip}

\vspace{\baselineskip}
\noindent \textbf{Reservoir computing mathematics.} RC can be modeled with technology-agnostic equations. 
Let $\mathcal{N}$ be the number of neurons in the reservoir, $\mathcal{R}<\mathcal{N}$ the number of readout neurons, $\mathcal{I}$ the number of inputs to the reservoir, and $\mathcal{J}$ the number of outputs from the RC.  The inputs to the reservoir can be denoted by $\mathbf{u}[t]$, a discrete time-varying vector of length $\mathcal{I}$.  The reservoir state can be denoted as $\mathbf{x}[t]$, also a discrete time-varying vector of length $\mathcal{N}$.  In the ESN model, $\mathbf{x}[t] = f\left(\mathbf{W}^{in}\mathbf{u}[t] + \mathbf{W}^{res}\mathbf{x}[t-1]\right)$, where $f$ is a non-linear activation function, $\mathbf{W}^{in}$ are the input weights, and $\mathbf{W}^{res}$ are the reservoir weights.  Note that at each time step, $\mathbf{x}[t]$ evolves according to a function of the current inputs and the previous reservoir state, specifying the reservoir's memory and ability to process temporal signals.

In this work, the trained output layer is linear to match the capabilities of MCAs.  Let $\mathcal{T}$ be the number of training points and $\mathbf{x}[t]$ be the states of the output neurons at time $t$.  Given a desired output vector $\mathbf{y}[t]$ of length $\mathcal{J}$, the $\mathcal{J} \times \mathcal{R}$ output weight matrix, $\mathbf{W}^{out} = \mathbf{YX}^\top (\mathbf{XX}^\top+\lambda \mathbf{I})^{-1}$, where $\mathbf{Y} = \left[\mathbf{y}[0],\; \mathbf{y}[1],\; ...,\; \mathbf{y}[\mathcal{T}-1]\right]$, $\mathbf{X} = \left[\mathbf{x}[0],\; \mathbf{x}[1],\; ...,\; \mathbf{x}[\mathcal{T}-1]\right]$, $\top$ denotes the transpose, $\lambda$ is the regularization factor, and $\mathbf{I}$ is the $\mathcal{N} \times \mathcal{N}$ identity matrix.  During reservoir operation, the RC binary output vector is computed as $\mathbf{\hat{y}}[t] = round\left(\mathbf{W}^{out}\mathbf{x}[t]\right)$, rounding to either $0$ or $1$.  During testing, the obtained RC outputs $\mathbf{\hat{y}}[t]$ are compared against the expected outputs $\mathbf{y}[t]$ to determine reservoir accuracy. 
    
\vspace{\baselineskip}
\noindent \textbf{Experimental methodology.} Electron beam lithography was performed using a Raith $50$ kV patterning tool on a Si substrate spin-coated with PMMA-$495$. After lithography, the substrate was developed in a MIBK:IPA ($1$:$3$) solution for $30$ seconds. A $13$ nm layer of Cobalt was deposited at a rate of $0.3$ \AA/s above a $7$ nm Ti adhesion layer using an e-beam evaporator at base pressure of $2*10^{-7}$ Torr. Finally, lift-off was performed by soaking the sample in hot Acetone for $30$ minutes.
Magnetic force microscopy was performed using a Bruker AFM system with a high-moment tip. The nominal resonant frequency of the cantilever, the lift height, and the scan rate were $70$ kHz, $80$ nm, and $0.2$ Hz respectively.

\vspace{\baselineskip}

\noindent \textbf{Micromagnetic simulation methodology.} Micromagnetic simulations were performed with mumax3 \cite{mumax3}.  Cylindrical nanomagnets (with a cell size of $2x2x2$ nm for the Waveform ID. and Boolean function evaluation tasks and $4x4x3$ nm for the ECA observer task) were simulated with the material parameters of CoFeB: saturation magnetization $M_{sat}=7.23*10^5$ A/m, exchange stiffness $A_{ex} = 1.3*10^{-11}$ J/m, Gilbert damping factor $\alpha = 0.05$, input anisotropy $Ku_i = 3.62*10^5$ J/$m^3$, and reservoir anisotropy $Ku_r = 1.05*10^5$ J/$m^3$.  Spatial and temporal parameters were chosen to maximize reservoir expressivity: nanomagnet diameter $d = 30$ nm, nanomagnet thickness $th = 12$ nm, and reservoir period $\tau \in [25, 30]$ ns. The reservoirs were designed through a combination of heuristic methods and annealing algorithms.

Inputs were provided to the reservoir by writing the magnetizations of the input nanomagnets to particular states; these input nanomagnets have a higher anisotropy, enhancing their ability to maintain their state after the writing force is removed.  In future experimental systems, this can be achieved by providing the input nanomagnets with greater interfacial anisotropy, greater thickness, or a continuous write current.

Input writing was considered to be instantaneous. This is justified by the fact that the reservoir relaxation time, $\tau \approx 30$ ns $ >> 3$ ns $ \approx \tau_{write}$ \cite{Lim16} is significantly slower than the write speeds available with modern MRAM writing techniques (STT or SOT) \cite{Lim16}. After relaxation during each cycle, the z-magnetization at the center of each nanomagnet is sampled and provided to the MCA output layer.

Simulations were performed at zero temperature, and it is expected that with tuning of reservoir geometry, damping, input frequency and amplitude, and output weights, the reservoir will operate similarly at non-zero temperatures as was experimentally demonstrated with nanomagnet logic systems \cite{Niemier_2011_NML}. The stray magnetic field from the MTJ reference layer is neglected, as compensating nanomagnets are conventionally included in the MTJ stack to counteract these effects \cite{CompNML_Liu14, CompNML_Shah14}. Perturbations of the nanomagnet state from the STT read process are similarly neglected, as any deviations in the MTJ resistance are expected to be consistent over time and are therefore incorporated into the MCA training. 





\vspace{\baselineskip}

\noindent{\textbf{Short-term memory and parity check capacity.}} STM and PC capacity are calculated as in \cite{STNOs_Kanao2019} according to $C = \sum_{i=0}^{\infty}corr^2(\mathbf{y_i}, \mathbf{\hat{y_i}})$, where $i$ is the delay of the STM and PC tasks with $\mathbf{y_{STM,i}}=\mathbf{u}[t-i]$ and $\mathbf{y_{PC,i}}=\bigoplus_{j=0}^{i}\mathbf{u}[t-j]$ where $\oplus$ is the binary sum or XOR operation.  For this work, both sums where taken to $i = 7$ as both sums converged very quickly and the correlation coefficients disappeared after $i = 5$.  The literature is inconsistent regarding whether the sum should begin from zero or one, creating a capacity differential of one.  This inconsistency has been adjusted for when comparing to the literature.

\vspace{\baselineskip}

\noindent \textbf{RC without a reservoir layer.} In order to prove that the NMRC is performing useful information processing, the RC results were compared to an RC with no reservoir layer. This network was trained to perform the tasks based on the inputs $u[t], u[t-1], ..., u[t-(m-1)]$ for some memory content $m$ that maximizes the testing accuracy. 

For the waveform identification task, $m$ is $5$. For the Boolean functions task, $m$ is $2$, $3$, and $4$ for $k = 2$, $3$, and $4$ respectively; this is intuitive as $k$ is the number of bits upon which the output, $y[t]$, is dependent.  For the ECA observer, increasing $m$ to an arbitrarily large value permits a feed-forward accuracy of $100\%$ for all $k$ due to the periodic nature of the task; therefore, to provide a fair comparison, the memory content in the comparison network was limited to that of the NMRC. As the memory content of the ECA observer NMRC was determined to be less than two bits, the time-multiplexed memory content, $m$, in the corresponding RC with no reservoir layer was limited to two bits.  

\vspace{\baselineskip}

\noindent \textbf{Area, energy, and delay of NMRC.}  The area, energy, and delay metrics of the NMRC are calculated in terms of the number of reservoir nanomagnets ($\mathcal{N}$), the number of output nanomagnets ($\mathcal{R}$), the number of input nanomagnets ($\mathcal{I}$), the length of the the output vector ($\mathcal{J}$), and the reservoir operating period ($\mathcal{\tau}$) -- which accounts for write, relaxation, and read delays as illustrated in Fig. \ref{fig:Figure2_NMRC}c.  As the output nanomagnets are a subset of the reservoir nanomagnets, there are a total of $\mathcal{N} + \mathcal{I}$ nanomagnets.  For energy estimates reported in this work, $\mathcal{R}$ was chosen as the minimum number of output nanomagnets that achieves within $0.5\%$ of the accuracy for $\mathcal{R} = \mathcal{N}$.  It is assumed that the feature size of the peripheral CMOS circuitry is $65$ nm. As the MCA weights need only be trained once upon initialization, this training cost is neglected.

\noindent \textit{\textbf{Area:}} The total area of the NMRC with input and output logic can be calculated as: $A_{total} = A_{NM} + A_{MCA} + A_{CMOS}$.  The area each nanomagnet occupies, including the spacing between nanomagnets, is approximately $0.0035$ $\mu m^2$, making the total reservoir area $A_{NM} = 0.0035$ $\mu m^2 * (\mathcal{N}+\mathcal{I})$.  Given an individual memristor area of $10F^2$, where $F$ is the feature size, $A_{MCA} = 10F^2 * \mathcal{N} * \mathcal{J} = 0.0423$ $\mu m^2 * \mathcal{N} * \mathcal{J}$ for a $65$ nm process.  As two PMOS and one NMOS transistor will be used for each output nanomagnet, and the area of each NMOS (PMOS) transistor is $4F^2$ ($8F^2$), $A_{CMOS} = 20F^2 * \mathcal{N} = 0.0845$ $\mu m^2 * \mathcal{N}$.  In total, $A_{total} = 0.0035 * (\mathcal{N}+\mathcal{I}) + 0.00423 * \mathcal{N} * \mathcal{J} + 0.0845 * \mathcal{N}$ $\mu m^2$.

It should be noted that it would be appropriate to vertically integrate nanomagnets, memristors, and CMOS in a three-dimensional heterogeneous stack. As this is prospective and has minimal impact on the comparison, this analysis considers only two-dimensional area.

\noindent \textit{\textbf{Energy:}} The energy cost per cycle for the NMRC is $E_{total} = E_{MTJwrite} + E_{MTJread} + E_{MCAread}$.  The power dissipated by a single MTJ during writing is $P_{MTJwrite\_single} = \frac{{V_{dd}}^2}{R_{avg}}$, where $Vdd$ is the supply voltage and $R_{avg}$ is the average resistance defined as: $R_{avg} = \frac{R_P + R_{AP}}{2}$,
where $R_P$ ($R_{AP}$) is the (anti-)parallel resistance of the MTJ.  The STT writing energy is therefore $E_{MTJwrite} = P_{MTJwrite\_single} * \mathcal{I} * \tau_{write}$, where $\tau_{write}$ is the write pulse length. For this work, $Vdd = 1.7$ V, $R_P = 25$ k$\Omega$, $R_{AP} = 35$ k$\Omega$, and $\tau_{write} = 3$ ns \cite{Lim16}. Thus, $E_{MTJwrite} = \frac{{V_{dd}}^2}{R_{avg}} * \mathcal{I} * \tau_{write} = 289$ fJ$ * I$.  The reference resistance will be $R_{AP}$, thus read energy through the voltage divider is $E_{MTJread} = \frac{{V_{dd}}^2}{R_{avg} + R_{AP}} * \mathcal{R} * \tau_{read} = 311$ fJ$ * \mathcal{R}$, assuming $\tau_{read} = 7$ ns. Memristor resistance must be significantly greater than $R_{AP}$ for proper voltage divider functionality. An average resistance of $1$ M$\Omega$ is assumed, which is within the standard range for memristors \cite{MCA_review_Li18}.  On average, the voltage across each memristor will be $Vdd/2$, applied for time $\tau_{read}$.  Thus, $E_{MCAread} = \frac{(Vdd/2)^2}{1 M\Omega}*\tau_{read} * \mathcal{R} * \mathcal{J} = 5.05$ fJ$ * \mathcal{R} * \mathcal{J}$, giving $E_{total} = (289 * \mathcal{I} + 311 * \mathcal{R} + 5.05 * \mathcal{R} * \mathcal{J})$ fJ.

\noindent \textit{\textbf{Delay:}} The delay of the NMRC is simply $\tau$: $D = \mathcal{\tau}$.

\vspace{\baselineskip}

\noindent \textbf{Area, energy, and delay for CMOS ESN.} The area, energy, and delay of a CMOS ESN was evaluated based on analysis of a design synthesized in a $65$ nm process with Cadence Genus Synthesis Suite.  The digital ESN system has a $16$-bit fixed-point vector-vector multiplier.  For a fair comparison between the NMRC and CMOS reservoirs, the output layers of both were implemented with MCAs.

The number of neurons in the synthesized ESN for each task was tuned to match the accuracy obtained with the NMRC.  For each task and number of neurons, ten samples of $20$ networks each were generated with random weights and evaluated on the task.  The accuracy of the best network in each sample was recorded as the sample accuracy, and these accuracies were averaged across the ten samples to give the reported accuracies in Fig \ref{fig:Figure5_Results}a.

\noindent \textit{\textbf{Area:}} The CMOS area was obtained directly after synthesis using the syn\_gen command in Genus.  The MCA area was determined similarly to the NMRC MCA. The total area is the sum of the CMOS and MCA areas.
 
\noindent \textit{\textbf{Energy:}} Energy was calculated using the total power dissipation reported by Genus (both static and dynamic) in concert with the input switching rates of random data.  To determine the energy, the power dissipation was multiplied by the delay per RC operation (calculated below).  The energy consumed by the MCA was calculated and added to the total, in the manner described above for NMRC.

\noindent \textit{\textbf{Delay:}}   Genus reported the longest path in the design, which was between $2$ ns and $3$ ns over the three tasks.  The maximum clock frequency is therefore between $300$ MHz and $500$ MHz. The minimum delay per operation is calculated by multiplying the minimum clock period ($T_{clk}$) by the number of clock cycles per reservoir operation: $D_{CMOS}=(\mathcal{N} + 6) * T_{clk}$.

\vspace{\baselineskip}

%

\bibliographystyle{IEEEtran}
\bibliography{IEEEabrv,main}

\vspace{1.5em}
{\large \noindent \textbf{Acknowledgements}}

\noindent Any opinions, findings and conclusions, or recommendations expressed in this material are those of the authors, and do not necessarily reflect the views of the US Government, the Department of Defense, or the Air Force Research Lab. Approved for Public Release; Distribution Unlimited: AFRL-2022-4420.  The authors thank E. Laws, J. McConnell, N. Nazir, L. Philoon, and C. Simmons for technical support, S. Luo for fruitful discussion, and the Texas Advanced Computing Center at The University of Texas at Austin, Austin, TX, USA, for providing computational resources.
F.G.S. acknowledges support from project No. PID2020117024GB-C41 funded by Ministerio de Ciencia e Innovacion from the Spanish government.

\end{document}